\pdfoutput=1

\documentclass[11pt]{article}

\usepackage[preprint]{acl}

\usepackage{times}
\usepackage{latexsym}

\usepackage[T1]{fontenc}

\usepackage[utf8]{inputenc}

\usepackage{microtype}

\usepackage{inconsolata}

\usepackage{graphicx}
\usepackage{paralist}
\usepackage{xcolor}
\usepackage{xspace}
\usepackage{amsmath}
\usepackage{cleveref}
\usepackage{tabularx}
\usepackage{multicol}
\usepackage{multirow}
\usepackage{booktabs}
\usepackage{enumitem}
\usepackage{amsmath}
\usepackage{amssymb}
\usepackage{mdframed}
\usepackage{longtable}

\makeatletter
\renewcommand\tabularxcolumn[1]{>{\@minipagetrue}p{#1}}
\makeatother

\newcommand{\dataset}{\textsc{PrimeX}\xspace}

\title{\dataset: A Dataset of Worldview, Opinion, and Explanation}

\author{  Rik Koncel-Kedziorski\textsuperscript{1},
  Brihi Joshi\textsuperscript{2},
  Tim Paek\textsuperscript{1} \\ 
   \textsuperscript{1}Apple,
  \textsuperscript{2}University of Southern California \\
    \small{
    \textbf{Correspondence:} \href{mailto:rikka@apple.com}{rikka@apple.com}}
 }

\begin{document}
\maketitle
\begin{abstract}
As the adoption of language models advances, so does the need to better represent individual users to the model. Are there aspects of an individual's belief system that a language model can utilize for improved alignment? Following prior research, we investigate this question in the domain of opinion prediction by developing \dataset, a dataset of public opinion survey data from 858 US residents with two additional sources of belief information: written explanations from the respondents for why they hold specific opinions, and the Primal World Belief survey for assessing respondent worldview. We provide an extensive initial analysis of our data and show the value of belief explanations and worldview for personalizing language models. Our results demonstrate how the additional belief information in \dataset can benefit both the NLP and psychological research communities, opening up avenues for further study. 
\end{abstract}

\section{Introduction}

Psychological research and clinical successes give evidence that a person's beliefs about themselves, their future, and their environment can significantly shape their behavior \citep{beck1976, Dweck1995ImplicitTA, Hofmann2012TheEO}. Recent work shows that an individual's {\it worldview} --- or beliefs about the overall character of the world --- can explain persistent behavioral patterns and correlates with personality, well-being, political, religious, and demographic variables \citep{clifton2019primalworldbeliefs}. As such, worldview can be seen as a powerful, compact, and predictive model of the individual's belief system. 

Simultaneously, advancements in language modeling have made it possible to incorporate higher-level user beliefs into predictive models \citep{sun2024personadbefficientlargelanguage}.
A better understanding of individual belief systems can improve personalization of language models (LMs), for instance by building better representations of an individual user's {\it persona} -- characteristics, preferences, and behavior. 
Persona-adapted language models (PA-LMs) have been used to create realistic simulated communities \cite{park2022socialsimulacracreatingpopulated, zhou2024sotopiainteractiveevaluationsocial,park2024generativeagentsimulations1000}, generate arbitrary amounts of diverse, synthetic data \cite{moon2024virtualpersonaslanguagemodels, ge2024scalingsyntheticdatacreation}, and simulate partners in training applications for a variety of professional domains~\citep{Markel2023GPTeachIT,Louie2024RoleplaydohED,10.1145/3613904.3642159}. 
A common evaluation of PA-LMs is predicting user responses to surveys and behavioral tests \cite{Argyle_2023, santurkar2023opinionslanguagemodelsreflect,hwang-etal-2024-graph, joshi2025improvingllmpersonasrationalization}.

\begin{figure*}[t]
    \centering
    \includegraphics[width=\linewidth]{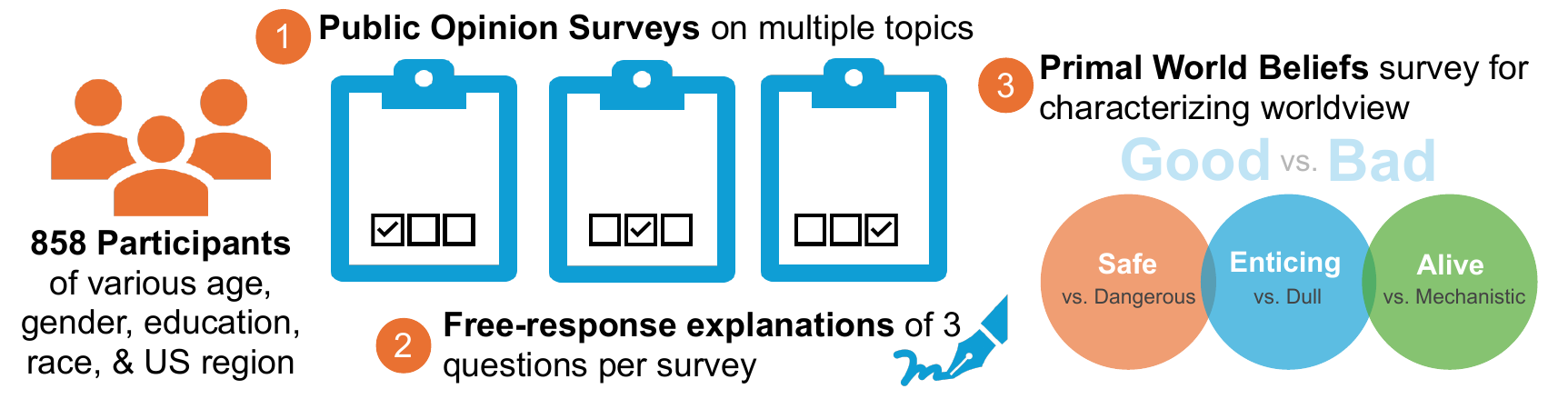}
    \caption{Overview of the \dataset data. We collect three types of responses from a diverse pool of participants: Opinions from 3 Pew Research surveys; explanations for 3 opinions per survey; and Primal World Belief survey of participant worldview.}
    \label{fig:overview}
\end{figure*}

To facilitate both worldview and persona research, we introduce the \dataset dataset of opinions, explanations, and beliefs about the world. \dataset consists of anonymous survey responses by 858 US residents from various geographic regions, age groups, education levels, and genders. Our respondents complete a subset of questions from each of three American Trends Panel public opinion surveys \citep{pew:ATP}. This enables the study of a single individual's opinions across different topics, which is not possible with existing datasets. We also collect two supplemental categories of user information which are, to our knowledge, novel in persona research. First, for a portion of opinion questions, we collect free-form written explanations from the respondent of why they hold a particular opinion. These explanations often draw on the respondent's higher-level beliefs about the world. 
We show that explanations improve a PA-LM's ability to predict an individual's other opinions.
Second, we collect participants' responses to the 18-question version of the Primal World Beliefs survey, an instrument for characterizing an individual’s worldview which generally takes less than 10 minutes to answer \citep{clifton2019primalworldbeliefs, Clifton2021BriefMO}. 

We find significant correlations between worldview and opinions across topics and show that worldview impacts stylistic characteristics of written explanations. Including worldview and explanations in user representations for PA-LMs can improve opinion prediction, and we develop an analysis of the utility of a given explanation to PA-LMs. Additionally, we show how an individual's Primal World Beliefs can be predicted from their opinions and explanations, an interesting new avenue for building general user representations from specific user data. 

Our experiments and analysis of \dataset data highlight the value of belief explanations and worldview for personalizing language models. Though extensive, they are far from exhaustive --- we believe this dataset constitutes a rich source of persona information for continued analysis in both the NLP and psychological research communities.\footnote{\url{https://github.com/apple/ml-primex}}

\section{Background}

\paragraph{Primal World Beliefs} Primal World Beliefs, or {\it Primals}, aim to capture an individual's beliefs about the general character of the world~\citep{clifton2019primalworldbeliefs}. Examples of Primals include \textit{The world is Safe} and \textit{The world is Interesting}. Research has shown these beliefs to be stable across time and correlated with a number of personality and well-being variables. 
We hypothesize that LMs have some knowledge of Primals due to how the theory of Primals itself was developed, which involved extensive linguistic analysis of text that is likely to be part of many LM's pretraining data. In particular, researchers scoured hundreds of historical texts (including sacred texts, novels, films, speeches, and philosophical works) and over 80k tweets for statements about how people view the world as a whole, using NLP extraction tools and Latent Dirichlet Allocation for topic clustering. Over a span of 5 years (2014-2019), they coded the statements and consulted with social science experts as well as religious focus groups to identify 26 Primal World Beliefs~\citep{clifton2019primalworldbeliefs}. 
Primals are organized under the top-level belief that \textit{The world is Good} and secondary beliefs that it is \textit{Safe} (versus dangerous), \textit{Alive} (intentionally and purposefully interacting with us versus inanimate and mechanical), and \textit{Enticing} (interesting and beautful versus dull and ugly). 
These beliefs were ultimately validated through multiple psychometric measures. 
In our dataset, participants filled out the 18-item survey ~\citep{Clifton2021BriefMO} measuring their top-level and secondary Primals as part of their 30 minute session; in general, it took less than 10 minutes for most participants to fill out the survey.

\paragraph{Public Opinion} Public opinion surveys are used in PA-LM research due to their easy availability and the rigorous validation of their construction over many decades~\citep{pew:ATP}. The complexity and deeply personal nature of opinion offers a difficult challenge for personalized ML, and only recently have models become powerful enough to take on this task. Prior opinion datasets for PA-LM research borrow data originally intended for demographic and economic analysis, resulting in limited individual information ~\citep{santurkar2023opinionslanguagemodelsreflect}. Our work enriches public opinion data by addressing several shortcomings that hinder generalization: 1) opinions from a single user across multiple topics are not available; and 2) demographic distributions of existing data can bias the output of LLMs, which often under-represent certain viewpoints. In addition, our data allows us to correlate opinion and demographics with new variables of interest: viz., worldview and explanation style. 

\paragraph{Explanations} Social scientists often conduct free-form interviews, in part because participant explanations of responses can provide deeper insights than structured formats~\cite{stanford2021americanvoices}.
Inspired by this, we ask our participants to explain a subset of their survey opinions in a free text format in hopes of deriving a better understanding of their personas.
A work similar to ours has demonstrated the value of conducting a free form interview followed by refinement processes, but this method of gathering persona information is both expensive and intensive for users~\citep{ge2024scalingsyntheticdatacreation}. Our work introduces a lower-cost persona format and elucidates how explanation interacts with both opinion and worldview.
Model-generated explanations of reasoning have proven useful for improving performance on many tasks~\citep{wei2023chainofthoughtpromptingelicitsreasoning}, including preference modeling and opinion prediction~\citep{sun2024personadbefficientlargelanguage,do-etal-2025-aligning, joshi2025improvingllmpersonasrationalization}.
We analyze human-written explanations and model-generated explanations for prediction, and identify characteristics of explanations which models can use to improve prediction.

\paragraph{Personalized LMs} The advancement of large language models has enabled new possibilities for personalized machine learning. Adapting an LM to the preferences of individuals can be done via alignment strategies such as RLHF and DPO, but these require expensive, large scale data \citep{Ouyang2022TrainingLM, Rafailov2023DirectPO, Wu2023FineGrainedHF}. Recent datasets for personalizing LMs address issues of representation~\citep{kirk2024prismalignmentdatasetparticipatory,aroyo2023dicesdatasetdiversityconversational}, but focus on demographics of human feedback data for conversational content. Weaker personalization can be accomplished quickly and cheaply using low-data techniques such as prompting \citep{hwang-etal-2023-aligning} or refinement \citep{sun2024personadbefficientlargelanguage}. 
These works make use of personas to adapt language models to an individual user's preferences. \dataset provides rich user data and can serve as training and testing data for personalization methods.

\paragraph{Psychology and LMs}
Popular psychological theories in NLP literature include Big 5 personality traits~\citep{goldberg1993bigfive} and Schwartz theory of basic values~\citep{shwartz1992test}. Recent works have analyzed the presence of personality traits in machine generated text \cite{Hilliard2024ElicitingPT}, and have even administered psychometric evaluations to various LMs to identify their default synthetic personality traits \cite{Karra2022AIPE,serapiogarcía2023personalitytraitslargelanguage}. Some works have incorporated psychological traits into PA-LMs \cite{moon2024virtualpersonaslanguagemodels, park2024generativeagentsimulations1000, li2025big5chat}.
Primal World Beliefs have been shown to explain broad aspects of personality~\citep{clifton2019primalworldbeliefs}, but remain underexplored in this space. To our knowledge, ours is the first work to collect worldview data for the purposes of analyzing and developing PA-LMs. 
\section{Dataset Construction}
The goal of \dataset is to extend current resources along multiple dimensions. Addressing a shortcoming in existing opinion data, we collect responses on multiple topics from each individual. This enables the development of personas which generalize across topics. For a subset of opinion questions, we collect free-form explanations for why the respondent holds their particular opinion. Explanations give insights into an individual's belief system and can also improve persona development. 
Lastly, we consider a source of user information which has not yet been brought to bear on PA-LM development: individual worldview. Hence, we collect responses to the 18 question Primal World Beliefs survey to capture an individual's worldview. 
In total, \dataset includes responses from 858 individuals, similar in size to recent works in LM personalization (\citet{park2024generativeagentsimulations1000}, $N=1052$) and personality psychology (\citet{Ludwig2022ResilienceOP}, $N=529$)

\subsection{Survey Questions}
Our data consists of three types of questions: public opinion, free-response explanations, and Primals. We use multiple-choice public opinion questions from the American Trends Panel surveys \citep{pew:ATP}, which have been carefully developed by experts at Pew Research to mitigate bias, ambiguity, difficulty, and other confounding factors. We use 3 surveys: ATP Wave 34, dealing with biomedical and food issues \citep{wave34}; ATP Wave 41, dealing with the condition of America in the year 2050 \citep{wave41}; and ATP Wave 54, dealing with economic inequality \citep{wave54}. From each of these surveys, we manually select 10 questions that meet two characteristics: they ask about personal opinions rather than biological or economic facts; and their response distribution in the larger population has higher entropy, as these are more likely to produce diverse answers from our participants. 
The full list of questions, response choices, and shorthand names used in this work (e.g. \textsc{Organic Foods}, \textsc{Govt Retirement}) are listed in Table~\ref{tab:pewquestions} in Appendix~\ref{sec:appendix_pew}.

\begin{figure}[!t]
\small
\begin{mdframed}[linecolor=black!30,backgroundcolor=black!5]
\textbf{Do you favor or oppose the use of animals in scientific research?}\\
\textbf{User 1:} Favor --- Most of the vaccines and oncology drugs were discovered and invented due to trials on animals which I think I favor \\
\textbf{User 2:} Oppose --- It is clear by now that animals experience a range of emotions just like we do, so what was once thought acceptable is no longer. Just because we have all the power doesn't mean we should inflict pain. \\
\textbf{User 3:} \ldots
\end{mdframed}

\caption{Examples of opinions with user explanations.}
\label{fig:explanation_examples}
\end{figure}

For 3 questions from each ATP survey, we ask participants to explain their answer in a free-response format. We instruct respondents to ``draw on any aspect of your personal history, social life, experiences, thoughts, feelings, beliefs, or values'' in their explanation (see Figure~\ref{fig:prompt_finetuned} for the full instructions). 
Examples of elicited explanations are shown in Figure~\ref{fig:explanation_examples} (additional examples in Appendix~\ref{sec:appendix_examples}).

We also include the 18 item Primal World Beliefs Inventory (PI-18)~\citep{Clifton2021BriefMO}. This shorter instrument balances brevity and granularity, measuring top-level ({\it Good}) and secondary Primals ({\it Safe}, {\it Enticing}, and {\it Alive}). The PI-18 has been shown to have high correlation with the full 99 question inventory. Questions in the PI-18 are multiple choice with responses ranging from "Strongly Disagree" to "Strongly Agree", which are converted to real-values ranging from 0 to 5. We follow the administration and scoring instructions given in \citet{primalAdmin} (questions and scoring functions are repeated in Appendix~\ref{sec:app-primals}).

In addition to the opinion, explanations, and Primal World Belief data, we also ask questions covering basic demographic self-identification: geographic region, age range, gender, English proficiency, number of children, employment status, political affiliation, hobbies, other languages spoken at home, and races. 

\subsection{Data Collection}
We recruit 858 US participants through a third-party user study firm, User Research International. Participants were selected to achieve relative balance in terms of male/female ratio, age range, and geographic distribution.\footnote{We provided a non-binary gender option, but we did not control for representative non-binary participation.} A condensed version of the demographic distribution of this data is provided in Table~\ref{table:demo_abridged} in Appendix~\ref{sec:app-demo}. Our respondents reflect a balance of geographic regions and age groups. To maintain this balance, we struggled to recruit male and female respondents at equal rates resulting in a female bias. Compared to the national average, people with college degrees or higher are overrepresented in our data. The reported race of our respondents shows nationally representative numbers of Black and Asian respondents, a slight over-representation of White respondents, and under-representation of Spanish, Hispanic, or Latino respondents. Future data collection efforts should consider additional controls for a more nationally representative demographic distribution if necessary. 

Each participant was offered payment above local minimum wage for their participation in the survey. The projected time to complete the survey was 30 minutes. Participants gave their informed consent before participation and were made aware of the intended uses of their data. They were offered the chance to stop at any time, and given the option to not answer any question. The survey design and collection process was guided by an institutional review board to ensure compliance with regulatory and ethical standards. After the survey, participants had a chance to review their responses and given the opportunity to opt-out and have their data removed from the dataset. We removed all of the data from participants who had opted-out.

\begin{table}
\footnotesize
    \begin{tabular}{l|ccccc|c}
    \toprule Primal & {\it N=} & Avg & Std & min & max & US Avg. \\ \hline
{\it Good} & 785  & 3.09  & 0.68  & 0.53  & 4.93 & 2.9 \\
{\it Safe} & 827  & 2.50 &  0.90  & 0.00  & 5.00 &  2.5 \\
{\it Alive} & 780  & 2.66 &  1.10  & 0.00  & 5.00 & 2.8 \\
{\it Enticing} & 830  & 3.73  & 0.75  & 0.57  & 5.00 & 3.7 \\
\bottomrule
    \end{tabular}
    \caption{Primal Belief scores of our respondents.}
\label{table:primal_stats}
\end{table}

The survey questions were presented in the same order for all participants: first the subsampled ATP Wave 34, ATP Wave 54, and ATP Wave 41 surveys with additional explanation questions, followed by the PI-18, and lastly some additional optional additional demographic questions. The 3 opinion questions which are each followed by explanations are given at the beginning of each section, followed by the 7 remaining opinion questions from the same ATP survey. This was done in an effort to reduce the cognitive load required by switching between topics. The order of questions within sections of the survey was also fixed.  
From each participant we collected 30 opinion question responses, 9 explanations, answers to the Primal World Beliefs survey comprising 18 scalar ranked questions, and 11 demographic attributes. 
\section{Analysis of Primals}
\label{sec:primalsopinions}

The PI-18 measures the top-level {\it Good} Primal and secondary {\it Safe}, {\it Enticing}, and {\it Alive} Primals. The aggregate statistics for our respondents is shown in Table~\ref{table:primal_stats} along with the US average reported in existing research~\citep{primalwebsite}. Our sample averages are similar to the US population, and our standard deviations cover the spread of reported scores. Our respondents can choose not to answer any questions including those needed to compute their score for a particular Primal, resulting in different but still large sample sizes $N$.

\subsection{Primals and Opinion}
\begin{table}
\centering
\small
\begin{tabular}{lc@{\hskip 0.2cm}cc@{\hskip 0.2cm}cc@{\hskip 0.2cm}cc@{\hskip 0.2cm}c}
\toprule
 & \multicolumn{2}{c}{\bf Good} & \multicolumn{2}{c}{\bf Safe} & \multicolumn{2}{c}{\bf Alive} & \multicolumn{2}{c}{\bf Enticing} \\ 
 & $\uparrow$ & $\downarrow$ & $\uparrow$ & $\downarrow$ &  $\uparrow$ & $\downarrow$ &  $\uparrow$ & $\downarrow$  \\ \hline
$+$ & 0.29 & 0.09 & 0.23 & 0.08 & 0.04 & 0.01 & 0.10 & 0.04 \\
$-$ & 0.25 & 0.56 & 0.19 & 0.43 & 0.62 & 0.71 & 0.31 & 0.48  \\
$\varnothing$ & 0.46 & 0.35 & 0.58 & 0.49 & 0.35 & 0.28 & 0.60 & 0.48 \\
\bottomrule
\end{tabular}
\caption{Primal Beliefs in the explanations from users with highest $\uparrow$ and lowest $\downarrow$ scores.  Rows correspond to evidence of high ($+$) or low ($-$) belief in the explanation, or no evidence ($\varnothing$)}
\label{table:primalPresence}
\end{table}
\begin{table}[t]
\small
\centering
\begin{tabular}{lccc}
\toprule
 Primal & $\uparrow$ & $\downarrow$ & $\Delta$ \\ \hline
{\it Good} & 171 & 183 & 6.84\% \\
{\it Safe*} & 154 & 187 & 21.43\% \\
{\it Alive*} & 160 & 254 & 58.91\% \\
{\it Enticing*} & 189 & 167 & -11.92\% \\
\bottomrule
\end{tabular}
\caption{Length of explanations from users with highest $\uparrow$ and lowest $\downarrow$ Primal scores. Rows marked * indicate significant differences at $p<0.01$.}
\label{table:primal_length}
\end{table}

We compute correlations between Primals and responses to each opinion question to determine the effect of a higher or lower score for each Primal on a person's opinion. We ignore ``Prefer not to answer'' opinion responses and respondents without a particular high-level Primal score on a per-question/primal basis, resulting in different but still sizable sample size $N$ for each correlation. Opinion questions with 2 response options are treated as binary; the remaining opinion questions are mapped to integers following \citet{santurkar2023opinionslanguagemodelsreflect}. 
We report Spearman's rank coefficient $\rho$ for ordinal (non-binary) and Pearson's correlation coefficient $r$ for all correlations. 
The full tables of correlations for all Primals and opinion questions are shown in Appendix~\ref{sec:full_correlations}.

\paragraph{Effect size} \citet{Funder2019EvaluatingES} recommend reporting effect sizes relative to a benchmark for comparison. One benchmark they suggest, which we use in this work, is the typical effect size found in a large scale literature review. The average effect size of 708 meta-analytically determined Pearson correlations in personality and individual difference research determined by \citet{Gignac2016EffectSG} is $r=0.19$. As such, we follow their suggested thresholds for relatively small ($r=0.1$), typical ($r=0.2$), and relatively large ($r=0.3$). 

\paragraph{Notable Effects}
We observe relatively large correlations between the opinion that children will have a better standard of living in the future (\textsc{Child Standard}) and high {\it Good} and {\it Safe} scores ($r=0.338$ and $r=0.326$ respectively). Not surprisingly, we observe a large correlation between the role of God versus evolution in determining the development of human life (\textsc{Evolution}) and high {\it Alive} scores ($r=0.322$). 
Interestingly, we also observe relatively large correlations between how much gas prices impact one's view of the economy (\textsc{Gas Prices}) and {\it Alive} scores ($r=0.310$). 

Overall, we observe small or stronger correlations ($r > 0.1$) between all Primals and at least some questions from each topic. The strongest correlations are found between Primals and questions from Wave 41 on the likely condition of America in 2050. This may indicate that Primals better encode how a person views the future world compared to the present one, but further analysis is needed. 

\subsection{Primals and Explanations}
Is a person's worldview evidenced in their explanations for their opinions? We test this using an LLM (zero-shot GPT-4o) to judge whether each explanation indicates the user has high ($+$) or low ($-$) Primal, or if there is no evidence ($\varnothing$). The judge is provided a definition of the Primal taken from \cite{primalwebsite}, shown in Figure~\ref{fig:analysis_rep}. We analyze the explanations from the 50 respondents with the highest ($\uparrow$) and lowest ($\downarrow$) scores for each Primal (900 explanations per Primal in total). Percentages of explanations with evidence for each group are shown in the columns of Table~\ref{table:primalPresence}. This analysis shows that indications of a respondent's Primals can be identified in their explanations. Across the $+$ and $-$ rows we see that indications of high (or low) primal belief appear more often in explanations from users with high (or low) scores for each Primal. Note that we intend this as a preliminary analysis. While the LLM-as-judge paradigm has been validated in some domains \citep{zheng2023judging}, additional work is needed to validate its use for predicting Primals. We explore the problem of predicting Primals by training models on \dataset in Section~\ref{sec:predictPrimals}.

As a more stable analysis, Table~\ref{table:primal_length} shows variations in length of explanations based on the respondent's Primal scores. The $\Delta$ column indicates the change moving from the high to low group for each Primal. We observe significant differences in average explanation length; respondents with lower {\it Alive} scores give over 50\% longer responses than their high {\it Alive} counterparts. 

An interpretable vocabulary analysis of explanations from these groups is complicated by the underlying topicality of the explanations, but predictive results of LMs conditioned on explanations in Section~\ref{sec:predictOpinions} shed some light on the differences between text from these groups. 
\section{Predicting User Responses}
\label{sec:predicting}
In order to highlight the value of \dataset for personalizing language models, we now consider the problem of predicting the survey responses of a user in our dataset using a PA-LM. Prior works on opinion prediction represent a user by their demographic attributes~\citep{santurkar2023opinionslanguagemodelsreflect} or by including a seed set of opinion questions and the user's answers~\citep{hwang-etal-2023-aligning}. 
We study the value of the additional data from \dataset --- Primals and explanations -- in user representations.
Our data also enables the analysis of representation generalization through the prediction of opinions of the same user across different topics. 
Finally, we explore whether a model can predict a user's Primals from their persona, and find that training on \dataset data facilitates this prediction.

\subsection{Opinion Prediction}
\label{sec:predictOpinions}
\begin{table}
\centering
\footnotesize
\begin{tabular}{lll}
\toprule
{\bf User Representation} & {\bf GPT-4o} & {\bf Mistral} \\ \hline
\multicolumn{3}{l}{\it All Topics}  \\ \hline
\textsc{Demographics} & 42.36 & 42.40 \\ 
\textsc{Demo \& Opinions} & 45.15 & 44.37 \\ 
\ + Primals & \underline{46.21} & 41.92 \\
\ + Explanations & \underline{48.12} & \underline{46.20}  \\ 
\ + Generated Explanations & \underline{46.30} & 45.13 \\ 
\textsc{\dataset persona} & \underline{48.31} & \underline{45.44}  \\\hline

\multicolumn{3}{l}{\it Cross Topic}  \\ \hline
\textsc{Demographics} & 39.24 & 39.72 \\ 
\textsc{Demo \& Opinions} & 39.82 & 40.12 \\
\ + Primals & 40.42 & \underline{41.03}  \\
\ + Explanations & 40.21 & 40.40  \\ 
\ + Generated Explanations & 40.02 & 40.37 \\ 
\textsc{\dataset persona} & \underline{40.68} & 40.78  \\
\bottomrule
\end{tabular}
\caption{Predicting user opinions from \dataset. Underlined results are significantly different from \textsc{Demo \& Opinions} from the same model at $p<0.05$. }
\label{table:gpt_pred_results}
\end{table}
\begin{table}
\centering
\footnotesize
\begin{tabular}{lcc}
\toprule
{\bf Primal} & {\bf Correlated} & {\bf Uncorrelated} \\ \hline

\textit{Good} & 51.07 & 40.05 \\ 
\textit{Safe} & 48.15 & 51.54 \\ 
\textit{Alive} & 54.88 & 41.07 \\ 
\textit{Enticing} & 53.27 & 44.89 \\ 

\bottomrule
\end{tabular}
\caption{Accuracy of model predictions for most and least correlated questions.}
\label{table:accByPrimal}
\end{table}

\paragraph{Task} In the opinion prediction task, a model is prompted with a user representation in text format and instructions to predict the user's response to unseen test questions one at a time. Test questions are provided in multiple-choice format. The general prompt format is shown in Figure~\ref{fig:prompt}. We report model accuracy, which is computed by comparing the user's true response token to a single token generated by the model.\footnote{We confirmed that the generated token always indicates an answer choice rather than reasoning or refusal.} See Appendix~\ref{sec:app-opiniontask} for additional details. 

We study two forms of this task: in the {\it All Topics} setting, test questions are drawn from all waves of the ATP survey. For user representations including seed opinions, we use the 9 explained opinions and test on the remaining 21 for each user. In the {\it Cross Topic} setting, seed opinions include 3 opinions with explanations and 7 additional opinions from Wave 34 and the 20 test questions come from Waves 41 and 54. This is a harder setting, since less is known about the user's opinion within a given test topic. To enable the study of finetuned models, we use a train/test split with 430 train and 428 test users. We explore the capability of both a large (GPT-4o~\citep{openai2024gpt4ocard}) and smaller (Mistral 7B Instruct~\citep{Jiang2023Mistral7}) for predicting survey responses from the \dataset test set given different user representations. 

\paragraph{User Representations} The main baseline for comparison is \textsc{Demo \& Opinions}, which represents users with their demographics and seed opinions. 
To this we add different types of novel data from \dataset: The {\it + Primals} setting includes information from the Primal World Beliefs survey. For long context models (GPT-4o), we include Primal scores with contextualizing information from~\citet{primalwebsite}; an example of this information is shown in Figure~\ref{fig:analysis_rep}. For short context models (Mistral) we provide the question/response pairs from the user's PI-18 (see Table~\ref{tab:primal_questions}).  The {\it + Explanations} setting includes the human written explanations for seed opinions (all 9 seed opinions have explanations in {\it All Topics}; 3 of the 10 in {\it Cross Topic} settings). The \textsc{\dataset persona} setting uses all information collected from users (demographics, seed opinions, explanations, and Primals). 

To explore the generalization capability of the explanations in \dataset, we study a {\it + Generated Explanations} setting. We use a finetuned GPT-4o to explain each seed opinion in the test set independently and include the generated explanations for each user in their representations. The explanation model is finetuned from GPT-4o with a user's demographics and a seed opinion as input and the user's explanation as the target output.  Finally, we include a demographics-only setting (\textsc{Demographics}) which allows the default alignment positions of models to be more prominent in the response distribution.  


\paragraph{Results} Table~\ref{table:gpt_pred_results} shows the average per-user zero-shot opinion prediction performance on the \dataset test set. Underlined results are significantly better than the \textsc{Demo \& Opinions} baseline for each model using paired t-tests with $p<0.05$. We see that both the explanations and worldview information provided in \dataset enables better prediction of unseen opinions. In the {\it All Topics} setting, GPT-4o can effectively use all information from \dataset including model-generated explanations, whereas Mistral requires human explanations to achieve significant results. In the {\it Cross Topic} setting, we see that both models struggle to generalize from off topic explanations. Mistral can use  worldview to make significant improvements in prediction, but GPT-4o requires both worldview and explanations to achieve statistically significant improvements. 

\paragraph{Primals and Accuracy}
Continuing the analysis of Section~\ref{sec:primalsopinions}, we compare the model prediction accuracy of five opinion questions from the {\it All Topics} test set which are most and least correlated with different Primals. 
Table~\ref{table:accByPrimal} shows the average accuracy of GPT-4o with \textsc{Demo \& Opinions}{\it ~+ Explanations} user representation for these questions. We see that for {\it Good}, {\it Alive}, and {\it Enticing} Primals the model is much better at predicting opinions for strongly correlated questions, but this is not true for {\it Safe world} belief. These trends hold for other user representations. 
This indicates that the Primal beliefs involved in the correlations identified in Section~\ref{sec:primalsopinions} are partly encoded by other user demographics, seed opinions, or explanations.
\begin{table}
\centering
\footnotesize
\begin{tabular}{lcccc}
\toprule
 & {\bf Good} & {\bf Safe} & {\bf Alive} & {\bf Enticing}\\ \hline
\multicolumn{5}{l}{\it GPT-4o} \\ \hline
\textsc{Demographics} & \underline{0.15} & \underline{0.13} & -0.04 & \underline{0.10} \\ 
\textsc{\dataset persona} & 0.06 & 0.05 & -0.07 & 0.03 \\ \hline
\multicolumn{5}{l}{\it Mistral} \\ \hline
\textsc{Demographics} & \underline{0.15} & \underline{0.11} & \underline{-0.12} & \underline{0.13} \\ 
\textsc{\dataset persona} & \underline{0.10} & 0.09 & -0.06& 0.06 \\ 
\bottomrule
\end{tabular}
\caption{Correlation of model accuracy and user Primal score. Underlined values are significant at $p<0.05$}
\label{table:accByUser}
\end{table}

In Table~\ref{table:accByUser} we show correlations between a user's Primals and opinion prediction accuracy under different user representations and models. Using the \textsc{Demographics} representation, models are more accurate for users with higher {\it Good}, {\it Safe} and {\it Enticing} beliefs, and for users with lower {\it Alive} beliefs. This aligns with results reported in \citet{santurkar2023opinionslanguagemodelsreflect} showing that the default values encoded in LLMs represent particular populations, but characterizes default values of LLMs in terms of worldview rather than demographic attributes. These correlations weaken in the \textsc{\dataset persona} setting, which also demonstrates better accuracy for opinion prediction. This indicates the additional user data in \dataset helps the LLMs to overcome their default values and personalize more effectively. 

\subsection{Primals Prediction}
\label{sec:predictPrimals}
\begin{table}[t]
\footnotesize
    \begin{tabular}{lcccc}
    \toprule  & {\bf Good} & {\bf Safe} & {\bf Alive} & {\bf Enticing} \\ \hline
    \textsc{Demo \& Opin.} & 0.56 & 1.13 & 1.46 & 0.61 \\
    \ {\it + Explanations} & 0.55 & 1.40 & 1.07 & 0.62 \\ 
    \textsc{Trained} & \underline{0.46} & \underline{0.69} & 1.22 & 0.65 \\
\bottomrule
    \end{tabular}
    \caption{Predicting a user's Primals (MSE). Underlined results are significantly better than both other methods at $p<0.05$} 
\label{table:predicting_primals}
\end{table}

In Section~\ref{sec:primalsopinions} we found evidence of Primals in written explanations. Can we predict a person's Primals based on their demographics, opinions, and explanations?
Table~\ref{table:predicting_primals} shows the performance of predicting Primal scores from various inputs, measured in mean squared error across test users. Here, the model is prompted with a persona description and tries to predict the user's responses to the PI-18. Scores for each Primal are computed from these synthesized responses and compared with the user's actual scores. The \textsc{Trained} predictor is a version of GPT-4o trained on the \dataset training data. It takes as input a user's demographics, seed opinions, and explanations and predicts the answer to each PI-18 item independently. 

The results in Table~\ref{table:predicting_primals} show varying degrees of success at recovering user Primals for the zero-shot \textsc{Demo \& Opinions} and {\it + Explanations} settings depending on the Primal being predicted. However, there is a significant improvement in predicting the top-level {\it Good} and secondary {\it Safe} scores using the \textsc{Trained} model; this suggests that it is possible for a model to learn to predict some aspects of a user's worldview from opinion and explanation data. It would be worth investigating if Primals can be approximated from other sources of user data.

\section{Measuring Utility of Explanations}
\label{sec:helpful}
Explanations have been shown to improve the predictive accuracy of PA-LMs overall, but do some explanations provide more information about a user's beliefs than others? What characteristics of an explanation allow a PA-LM to learn generalizable information about the user? 

To study this, we define a utility function $\mathcal{M}$ which expresses how useful a user's explanation of a seed opinion is to the PA-LM when predicting test opinions. We model this as the expected log-likelihood gain in the true class when predicting a user's opinions conditioned on the explanation compared to a baseline.\footnote{This is equivalent to the reduction in cross-entropy loss.} For the baseline, we condition the LM on a simple user representation consisting of demographic information plus a single unexplained seed opinion. See Appendix~\ref{sec:app_utility} for the full details of how we compute $\mathcal{M}$. 
Note that $\mathcal{M}$ may be negative, as some explanations cause the language model to move probability mass away from the user's test responses. 

\begin{figure}[t]
    \centering
    \includegraphics[width=\linewidth]{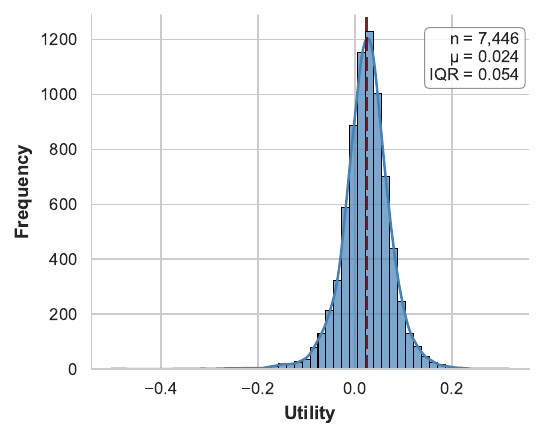}
    \caption{Distribution of Utility Scores for Explanations in \dataset.}
    \label{fig:util_dist}
\end{figure}

We compute $\mathcal{M}$ for explanations of answered seed opinion questions. The distribution of scores is shown in Figure~\ref{fig:util_dist}. Scores on this dataset range from -0.503 to 0.317 with a mean of 0.024. This is a non-normal distribution, so we report the interquartile range (IQR) of scores as a measure of variability. The IQR for utility of \dataset explanations is 0.054, indicating that most scores are clustered close to the mean. 

\begin{table}
\small
\centering
\begin{tabular}{lccccc}
\toprule
 & {\bf Length} & {\bf Good} & {\bf Safe} &  {\bf Alive} &  {\bf Enticing}\\ \hline
 High  & 333 & 0.58 & 0.44 & 0.59 & 0.36 \\
 Low  & 144 & 0.61 & 0.53 & 0.69 & 0.53 \\
\bottomrule
\end{tabular}
\caption{Length and evidence of Primals in high and low utility explanations.}
\label{table:bestworst}
\end{table}

\paragraph{Quantitative Characterization}
To characterize the textual difference between the highest and lowest utility explanations, we group the explanations with utility values beyond $1.5 \times$ IQR from the first and third quartiles. In Table \ref{table:bestworst} we repeat the LLM-as-judge experiment introduced in Section~\ref{sec:primalsopinions}. Here we report the percentage of explanations which indicate any valence of the Primal (i.e. $+$ and $-$ versus $\varnothing$). Notably, the low utility explanations are marked as more indicative of a user's worldview across Primals. Developing PA-LMs that can utilize these worldview-revealing texts may be a fruitful avenue for future work. 

\begin{table}[t]
\footnotesize
\centering
\begin{tabular}{l|cc}
\toprule
 & {\bf Seed Question} & {\bf Test Questions} \\ \hline
 High & 0.609 & 0.182 \\
 Low & 0.484 & 0.151\\
\bottomrule
\end{tabular}
\caption{High and low utility explanation similarity with seed and test question/response pairs.}
\label{table:bestsimilarity}
\end{table}
We also see that high utility explanations are on average over twice as long as low utility ones. Longer explanations may include more overlapping information with test questions and provide additional signal to the PA-LM for predicting user responses to these.
To test this, we measure the average cosine similarity of user explanation with the test questions and user responses.\footnote{Texts are embedded using the \texttt{all-MiniLM-L6-v2} model~\citep{reimers-2019-sentence-bert}.} Table~\ref{table:bestsimilarity} shows that the highest utility explanations are more similar to both the seed questions they explain and the user's test set. 



\paragraph{Effect of Question on Explanation Utility} 

\begin{figure}[t]
    \centering
    \includegraphics[width=\linewidth]{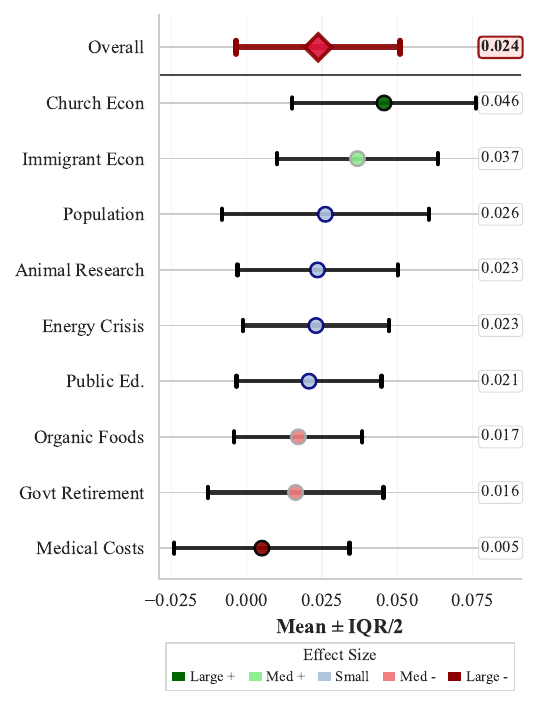}
    \caption{Utility Scores by Question. This figure is best viewed in color.}
    \label{fig:util_q}
\end{figure}

Certain questions in \dataset seem to elicit higher utility explanations. A plot of explanation utility by question is shown in Figure~\ref{fig:util_q}. We compare the distribution of utility scores for each question against the average for all questions using Welch's T-test and report effect size in the figure. The question eliciting the highest utility explanations is \textsc{Church Econ}; the lowest utility explanations are in response to \textsc{Medical Costs}. On average, each question yields explanations with positive utility, but there is large variance within groups. Based on these results, future work should consider the expected utility of the collected explanations when crafting elicitation materials. 
\section{Conclusion}
We introduce \dataset, a novel dataset of opinion question responses, explanations from respondents, and their answers to the Primal World Beliefs survey. We provide new insights into the relationships between personal opinions and worldview, and conduct a detailed analysis of the utility of user beliefs in PA-LMs. The analyses described here are only some of what is possible with \dataset. 

\dataset is the first dataset with opinions, explanations, and worldview data from the same individuals and can facilitate the development of methods for realistic LLM-based simulation. Importantly, \dataset allows for the modeling of user behavior based on individual beliefs rather than demographic attributes. Primal World Beliefs are validated constructs with strong explanatory power for a range of human behaviors. The quality of a simulated persona with a given set of Primals can be assessed based on the similarity of its opinions and explanations to comparable respondents in \dataset. Likewise, the quality of a simulated explanations can be improved by considering the \dataset data. In personality psychology, \dataset can be combined with existing data for continued development of the theory of Primal World Beliefs; it can also enable further analysis, including connections between demographics and worldview, qualitative coding of explanations, or the study of groups based on shared opinions.

In general, we hope that \dataset will inspire future research modeling not only user behavior but underlying user beliefs, as we observe that these correlate with the behaviors studied here. We encourage the NLP and psychological research communities to make use of this resource. 
\section{Limitations}
The participant pool for \dataset was restricted to English-speaking US. residents. We faced challenges collecting data from all demographic groups either equally or in proportion to that group's portion of the United States' population. As a result, \dataset under-represents ``Spanish, Hispanic, or Latino'' respondents and ``Male'' respondents. Due to the cost of collecting survey data, the number of participants in \dataset is relatively small for the purposes of training NLP systems. The online format of this survey may have posed additional problems for people with less technological familiarity. Particularly, if a respondent did not have access to text-to-speech on their device they would have had to type out their explanation answers, a burden for those with weaker typing skills. This could have resulted in suboptimal collection of their explanations. 

This work uses GPT-4o (gpt-4o-2024-11-20) accessed through the OpenAI API. This model is subject to a proprietary license which may change. The specific model may not be available indefinitely which impacts the reproducibility of the results reported in this paper. We also use Mistral 7B Instruct (v0.3), which is subject to the Apache 2.0 license. 

\section{Ethical Considerations}
The intention of \dataset is to provide researchers from the psychological and NLP research science communities a rich source of data for analysis of opinion, explanation, and worldview. Our data contains subjective opinions from respondents which may be offensive to some people. Our data was collected under the guidance of an internal review board to ensure participant safety. Participants gave their informed consent before participating in the survey. Participants had the option to refuse to answer any question. After completing the survey, participants had the option to withdraw their responses from the data release.

We study the impact of richer persona information for prompting LMs on the assumption that better user representations will enable more positive user experiences.  Language models and especially PA-LMs have been shown to exhibit unfair biases \citep{gupta2024biasrunsdeepimplicit}. We believe that richer user representations can counteract these biases by encouraging models to consider the individuality of each user rather than resorting to coarse generalizations.

\bibliography{custom}

\appendix

\section{Data Details}
\subsection{Demographics}
\label{sec:app-demo}
\begin{table}
\small
\centering
\begin{tabular}{lr}
    \toprule
      & {\it US Region} \\ \hline
 South  &  269  \\
 West  &  263  \\
 Midwest  &  183  \\
 Northeast  &  143  \\ \hline
 &  {\it Age} \\ \hline
30 to 49  &  231  \\
50 to 64  &  214  \\
65 or older  &  212  \\
18 to 29  &  200  \\ \hline
 &  {\it Gender} \\ \hline
Female  &  489  \\
Male  &  348  \\
Non-binary  &  18  \\
Prefer not to say  &  3  \\ \hline
 &  {\it Education} \\ \hline
High school  &  311  \\
Undergraduate degree (Bachelor's)  &  271  \\
Graduate or Professional degree  &  142  \\
Associate's degree  &  130  \\
Other responses  &  4  \\ \hline
 &  {\it Race}\\ \hline
White or Caucasian  &  554  \\
Black or African American  &  122  \\
Asian  &  79  \\
Spanish, Hispanic, or Latino  &  65  \\
Other responses  &  38  \\ \bottomrule
    \end{tabular}
    \caption{Demographic distribution of \dataset.}
\label{table:demo_abridged}
\end{table}

Demographic distribution is shown in Table~\ref{table:demo_abridged}

\subsection{Pew Survey Questions}
\label{sec:appendix_pew}

Table~\ref{tab:pewquestions} lists the public opinion questions in \dataset. All questions are taken from Pew Survey website. An option ``Prefer not to answer'' was included for all multiple choice questions to meet internal review requirements. We made slight formatting changes compared to Pew presentation to accommodate our survey software. 

\clearpage
{\onecolumn
\footnotesize
\begin{longtable}{p{0.15\linewidth}p{0.35\linewidth}p{0.40\linewidth}}
\caption{Public opinion survey questions in \dataset. For questions marked with \dag~we elicit explanations of participant responses.}\label{tab:pewquestions}\\

\toprule
\textbf{Name} & \textbf{Question Text} & \textbf{~ ~ Options} \\ 
\midrule
\endfirsthead

\multicolumn{3}{c}%
{{\bfseries \tablename\ \thetable{} -- continued from previous page}}\\
\toprule
\textbf{Name} & \textbf{Question Text} & \textbf{~ ~ Options} \\ \midrule
\endhead

\midrule
\multicolumn{2}{r}{{Continued on next page}} \\
\midrule
\endfoot

\bottomrule
\endlastfoot
\hline 
& & \multicolumn{1}{r}{\it Wave 34}  \\ \hline 
\dag Medical Costs & Which of these statements comes closer to your point of view, even if neither is exactly right? & \begin{compactenum}  
\item Medical treatments these days often create as many problems as they solve
\item Medical treatments these days are worth the costs because they allow people to live longer and better quality lives
\end{compactenum} \\
\hline \dag Animal Research & All in all, do you favor or oppose the use of animals in scientific research? & \begin{compactenum}  
\item Oppose
\item Favor
\end{compactenum} \\
\hline \dag Organic Foods & Do you think organic fruits and vegetables are generally… & \begin{compactenum}  
\item Worse for one's health than conventionally grown foods
\item Neither better nor worse for one's health than conventionally grown foods
\item Better for one's health than conventionally grown foods
\end{compactenum} \\
\hline Gene Risks & Thinking about what you have heard or read, how well do you think medical researchers understand the health risks and benefits of changing a baby's genetic characteristics? & \begin{compactenum}  
\item Not well at all
\item Not too well
\item Fairly well
\item Very well
\end{compactenum} \\
\hline Gene Disease & Do you think changing a baby's genetic characteristics to treat a serious disease or condition the baby would have at birth is an appropriate use of medical technology? & \begin{compactenum}  
\item Taking medical technology too far
\item An appropriate use of medical technology
\end{compactenum} \\
\hline Meat Hormone & How much health risk, if any, does eating meat from animals that have been given antibiotics or hormones have for the average person over the course of their lifetime? & \begin{compactenum}  
\item No health risk at all
\item Not too much health risk
\item Some health risk
\item A great deal of health risk
\end{compactenum} \\
\hline New Treatments & Thinking about medical treatments these days, how much of a problem, if at all, is the following:   New treatments are made available before we fully understand how they affect people's health & \begin{compactenum}  
\item Not a problem
\item A small problem
\item A big problem
\end{compactenum} \\
\hline Science Funding & Which statement comes closer to your view, even if neither is exactly right? & \begin{compactenum}  
\item Private investment will ensure that enough scientific progress is made, even without government investment
\item Government investment in research is ESSENTIAL for scientific progress
\end{compactenum} \\
\hline Food Additives & Which of these statements comes closer to your view, even if neither is exactly right? & \begin{compactenum}  
\item The average person is exposed to additives in the food they eat every day but they eat such a small amount that this does not pose a serious health risk
\item The average person is exposed to additives in the food they eat every day, which pose a serious risk to their health
\end{compactenum} \\
\hline Evolution & Thinking about the development of human life on Earth: Which statement comes closest to your view? & \begin{compactenum}  
\item Humans have evolved over time due to processes that were guided or allowed by God or a higher power
\item Humans have existed in their present form since the beginning of time
\item Humans have evolved over time due to processes such as natural selection; God or a higher power had no role in this process
\end{compactenum} \\
\hline & & \multicolumn{1}{r}{\it Wave 54}  \\ 
\hline \dag Govt Retirement & Do you think adequate income in retirement is something the federal government has a responsibility to provide for all Americans? & \begin{compactenum}  
\item No, not the responsibility of the federal government to provide
\item Yes, a responsibility of the federal government to provide for all Americans
\end{compactenum} \\
\hline \dag Church Econ & How much responsibility, if any, should churches and other religious organizations have in reducing economic inequality in our country & \begin{compactenum}  
\item None
\item Only a little
\item Some
\item A lot
\end{compactenum} \\
\hline \dag Immigrant Econ & How much, if at all, do you think the growing number of legal immigrants working in the US contributes to economic inequality in this country? & \begin{compactenum}  
\item Contributes not at all
\item Contributes not too much
\item Contributes a fair amount
\item Contributes a great deal
\end{compactenum} \\
\hline Gas Prices & How much, if at all, do you think gas prices are contributing to your opinion about how the economy is doing? & \begin{compactenum}  
\item Not at all
\item Not too much
\item A fair amount
\item A great deal
\end{compactenum} \\
\hline House Prices & How much, if at all, do you think real estate values are contributing to your opinion about how the economy is doing? & \begin{compactenum}  
\item Not at all
\item Not too much
\item A fair amount
\item A great deal
\end{compactenum} \\
\hline Job Confidence & How much, if at all, do you think the availability of jobs in your area are contributing to your opinion about how the economy is doing? & \begin{compactenum}  
\item Not at all
\item Not too much
\item A fair amount
\item A great deal
\end{compactenum} \\
\hline Race Econ & How much, if at all, do you think discrimination against racial and ethnic minorities contributes to economic inequality in this country? & \begin{compactenum}  
\item Contributes not at all
\item Contributes not too much
\item Contributes a fair amount
\item Contributes a great deal
\end{compactenum} \\
\hline Corporate Econ & How much, if at all, do you think regulation of major corporations contributes to economic inequality in this country? & \begin{compactenum}  
\item Contributes not at all
\item Contributes not too much
\item Contributes a fair amount
\item Contributes a great deal
\end{compactenum} \\
\hline Benefits Econ & How much, if at all, do you think the following proposals would do to reduce economic inequality in the U.S.?    Expanding government benefits for the poor & \begin{compactenum}  
\item Nothing at all
\item Not too much
\item A fair amount
\item A great deal
\end{compactenum} \\
\hline Antitrust Econ & How much, if at all, do you think the following proposals would do to reduce economic inequality in the U.S.?    Breaking up large corporations & \begin{compactenum}  
\item Nothing at all
\item Not too much
\item A fair amount
\item A great deal
\end{compactenum} \\
\hline & & \multicolumn{1}{r}{\it Wave 41}  \\ \hline \dag Population & In 2050, do you think population growth in the US will be a ... & \begin{compactenum}  
\item Not a problem
\item Minor problem
\item Major problem
\end{compactenum} \\
\hline \dag Energy Crisis & How likely do you think it is that the following will happen in the next 30 years? The world will face a major energy crisis & \begin{compactenum}  
\item Will definitely not happen
\item Will probably not happen
\item Will probably happen
\item Will definitely happen
\end{compactenum} \\
\hline \dag Public Ed. & Thinking ahead 30 years, which do you think is more likely to happen in the U.S.? & \begin{compactenum}  
\item The public education system will get worse
\item The public education system will improve
\end{compactenum} \\
\hline Child Standard & Thinking ahead 30 years, which do you think is more likely to happen in the U.S.? & \begin{compactenum}  
\item Children will have a worse standard of living
\item Children will have a better standard of living
\end{compactenum} \\
\hline China vs US & How likely do you think it is that the following will happen in the next 30 years? China will overtake the US as the world's main superpower & \begin{compactenum}  
\item Will definitely not happen
\item Will probably not happen
\item Will probably happen
\item Will definitely happen
\end{compactenum} \\
\hline Race Relations & Thinking ahead 30 years, which do you think is more likely to happen in the U.S.? & \begin{compactenum}  
\item Race relations will improve
\item Race relations will get worse
\end{compactenum} \\
\hline Climate Change & Thinking about the future of our country, how worried are you, if at all, about climate change? & \begin{compactenum}  
\item Not worried at all
\item Not too worried
\item Fairly worried
\item Very worried
\end{compactenum} \\
\hline Alzheimer Cure & How likely do you think it is that the following will happen in the next 30 years? There will be a cure for Alzheimer's disease & \begin{compactenum}  
\item Will definitely not happen
\item Will probably not happen
\item Will probably happen
\item Will definitely happen
\end{compactenum} \\
\hline Military Cost & If you were deciding what the federal government should do to improve the quality of life for future generations, what priority would you give to reducing military spending? & \begin{compactenum}  
\item Should not be done
\item A lower priority
\item An important, but not a top priority
\item A top priority
\end{compactenum} \\
\hline Religion & Thinking ahead 30 years, which do you think is more likely to happen in the U.S.? & \begin{compactenum}  
\item Religion will be about as important as it is now
\item Religion will become less important
\end{compactenum}\\
\bottomrule
\end{longtable}
}
\clearpage
\subsection{Explanation Examples}
\label{sec:appendix_examples}
\begin{figure*}[!t]
\small
\begin{mdframed}[linecolor=black!30,backgroundcolor=black!5]
\textbf{How much responsibility, if any, should churches and other religious organizations have in reducing economic inequality in our country?}\\
\textbf{User 11:} Only a little --- In my opinion, church members should address social and economic issues only as expressions of their faith. Other than that, there should be strict separation of church and state. \\
\textbf{User 12:} None --- Many religions teach the importance of charity, but in a country with no state official religion, we should not depend on, or demand, some or all religious organizations be part of a nationwide effort to redistribute wealth. Extremely large organizations, such as megachurches, should become taxable to an extent, but as a society, we should use our framework of government to reduce economic inequality, not attempt to create a system based on vastly different religions working together.  \\
\textbf{User 13:} A lot --- Churches are social groups. We should support ourselves as a community and churches are part of the community .
\ldots
\\ \\
\textbf{In 2050, do you think population growth in the US will be a ...}\\
\textbf{User 234:} Major Problem --- We are growing really fast. I know all over the world and the US, we don't have enough for people. That includes basics and I know growth is just going up still.  \\
\textbf{User 235:} Not a problem --- It will be opposite, population will be less than they expect given no one is having babies these days  \\
\textbf{User 236:} Minor Problem --- I don't expect population growth to be unmanageable if we do a good job managing it. The US is a huge and vast country with more than enough room and resources to handle population growth, especially if it lets more cities outside of the main urban areas grow.
\ldots \\ \\ 
\textbf{Do you think organic fruits and vegetables are generally ...}\\
\textbf{User 58:} Better for one's health than conventionally grown foods --- Fruits n vegetables are way more better than supplements and medicines \\
\textbf{User 59:} Neither better nor worse for one's health than conventionally grown foods --- I do not ever consume organic products because there are no legal or official standards for organic farming practices, although I do not believe those foods are necessarily worse than non-organic foods. I simply think those foods are marked up unnecesarrily to take advantage of a recent trend, even though those products are often inferior (smaller, less hearty, more prone to disease, etc). \\
\textbf{User 60:} Better for one's health than conventionally grown foods --- I've read a lot of research on the dangers of consuming pesticides. Pesticides are toxic to humans as well as other important life like pollinating insects. 
\ldots \\ \\ 
\end{mdframed}
\caption{Examples of opinions with user explanations.}
\label{fig:add_expl_examples}
\end{figure*}

Figure~\ref{fig:add_expl_examples} provides additional examples of explanations from \dataset.

\section{Full Correlations}
\label{sec:full_correlations}
This section contains all correlation results between Primals and survey responses.
\begin{table*}[ht]
\footnotesize
\centering
\begin{tabular}{lcrrrr}
\toprule
\multicolumn{6}{c}{\textbf{Good}} \\
\textbf{Question} & \textit{n} & $r$ & $p$ of $r$ & $\rho$ & $p$ of $\rho$ \\ \midrule
Medical Costs & 770 & 0.24 & 1.39e-11 & -- & -- \\
Animal Research & 732 & 0.108 & 3.42e-03 & -- & -- \\
Organic Foods & 785 & 0.076 & 3.36e-02 & 0.078 & 2.85e-02 \\
Gene Risks & 754 & 0.103 & 4.57e-03 & 0.103 & 4.75e-03 \\
Gene Disease & 736 & 0.082 & 2.54e-02 & -- & -- \\
Meat Hormone & 776 & 0.024 & 5.12e-01 & 0.027 & 4.50e-01 \\
New Treatments & 781 & -0.032 & 3.73e-01 & -0.018 & 6.18e-01 \\
Science Funding & 764 & 0.058 & 1.10e-01 & -- & -- \\
Food Additives & 778 & -0.042 & 2.48e-01 & -- & -- \\
Evolution & 785 & 0.037 & 3.01e-01 & -- & -- \\
Govt Retirement & 785 & -0.098 & 5.83e-03 & -- & -- \\
Church Econ & 768 & 0.116 & 1.30e-03 & 0.121 & 7.58e-04 \\
Immigrant Econ & 759 & -0.095 & 9.18e-03 & -0.102 & 5.11e-03 \\
Gas Prices & 780 & 0.043 & 2.35e-01 & 0.051 & 1.52e-01 \\
House Prices & 782 & -0.028 & 4.41e-01 & -0.009 & 7.92e-01 \\
Job Confidence & 782 & -0.01 & 7.88e-01 & 0.006 & 8.64e-01 \\
Race Econ & 775 & -0.023 & 5.31e-01 & -0.031 & 3.90e-01 \\
Corporate Econ & 768 & -0.122 & 7.21e-04 & -0.119 & 9.55e-04 \\
Benefits Econ & 779 & -0.039 & 2.80e-01 & -0.048 & 1.79e-01 \\
Antitrust Econ & 777 & -0.144 & 5.85e-05 & -0.154 & 1.66e-05 \\
Population & 773 & -0.169 & 2.30e-06 & -0.184 & 2.58e-07 \\
Energy Crisis & 762 & -0.063 & 8.42e-02 & -0.062 & 8.79e-02 \\
Public Ed. & 741 & 0.286 & 2.05e-15 & -- & -- \\
Child Standard & 741 & 0.338 & 3.27e-21 & -- & -- \\
China vs US & 766 & -0.178 & 7.42e-07 & -0.173 & 1.45e-06 \\
Race Relations & 751 & -0.274 & 1.96e-14 & -- & -- \\
Climate Change & 783 & -0.02 & 5.70e-01 & -0.008 & 8.26e-01 \\
Alzheimer Cure & 779 & 0.183 & 2.63e-07 & 0.181 & 3.91e-07 \\
Military Cost & 781 & -0.047 & 1.88e-01 & -0.046 & 2.03e-01 \\
Religion & 770 & -0.125 & 5.35e-04 & -- & -- \\
\bottomrule
\end{tabular}
\caption{Correlations of Pew Opinion responses with Good primal. Table shows number of responses \textit{n}, Pearson correlation coefficient $r$ with p-value $p$ of $r$, Spearman rank correlation coefficient $\rho$ with p-value $p$ of $\rho$. For questions with only 2 answer options, Spearman rank correlation is unavailable.}
\label{table:full_good}
\end{table*}

\begin{table*}
\centering
\footnotesize
\begin{tabular}{lcrrrr}
\toprule
\multicolumn{6}{c}{\textbf{Safe}} \\
\textbf{Question} & \textit{n} & $r$ & $p$ of $r$ & $\rho$ & $p$ of $\rho$ \\ \midrule 
Medical Costs & 806 & 0.259 & 8.62e-14 & -- & -- \\
Animal Research & 768 & 0.144 & 6.43e-05 & -- & -- \\
Organic Foods & 827 & 0.029 & 4.00e-01 & 0.031 & 3.80e-01 \\
Gene Risks & 787 & 0.087 & 1.42e-02 & 0.076 & 3.26e-02 \\
Gene Disease & 767 & 0.053 & 1.40e-01 & -- & -- \\
Meat Hormone & 815 & -0.128 & 2.59e-04 & -0.125 & 3.64e-04 \\
New Treatments & 822 & -0.132 & 1.41e-04 & -0.122 & 4.40e-04 \\
Science Funding & 801 & 0.051 & 1.52e-01 & -- & -- \\
Food Additives & 817 & -0.144 & 3.68e-05 & -- & -- \\
Evolution & 827 & 0.012 & 7.24e-01 & -- & -- \\
Govt Retirement & 827 & -0.129 & 2.02e-04 & -- & -- \\
Church Econ & 802 & 0.058 & 1.02e-01 & 0.059 & 9.63e-02 \\
Immigrant Econ & 798 & -0.088 & 1.25e-02 & -0.075 & 3.43e-02 \\
Gas Prices & 822 & -0.036 & 3.02e-01 & -0.031 & 3.70e-01 \\
House Prices & 820 & -0.108 & 1.96e-03 & -0.099 & 4.43e-03 \\
Job Confidence & 823 & -0.056 & 1.10e-01 & -0.057 & 9.95e-02 \\
Race Econ & 815 & -0.069 & 5.02e-02 & -0.075 & 3.16e-02 \\
Corporate Econ & 805 & -0.184 & 1.35e-07 & -0.188 & 7.75e-08 \\
Benefits Econ & 816 & -0.082 & 1.94e-02 & -0.104 & 2.89e-03 \\
Antitrust Econ & 814 & -0.166 & 2.06e-06 & -0.177 & 3.73e-07 \\
Population & 810 & -0.193 & 3.16e-08 & -0.206 & 3.29e-09 \\
Energy Crisis & 798 & -0.132 & 1.83e-04 & -0.146 & 3.63e-05 \\
Public Ed. & 774 & 0.284 & 7.10e-16 & -- & -- \\
Child Standard & 766 & 0.326 & 1.84e-20 & -- & -- \\
China vs US & 796 & -0.177 & 4.89e-07 & -0.174 & 8.41e-07 \\
Race Relations & 784 & -0.289 & 1.49e-16 & -- & -- \\
Climate Change & 824 & -0.04 & 2.53e-01 & -0.04 & 2.57e-01 \\
Alzheimer Cure & 816 & 0.067 & 5.72e-02 & 0.069 & 4.80e-02 \\
Military Cost & 820 & -0.037 & 2.92e-01 & -0.033 & 3.46e-01 \\
Religion & 805 & -0.09 & 1.10e-02 & -- & -- \\
\bottomrule
\end{tabular}
\caption{Correlations of Pew Opinion responses with Safe primal. Table shows number of responses \textit{n}, Pearson correlation coefficient $r$ with p-value $p$ of $r$, Spearman rank correlation coefficient $\rho$ with p-value $p$ of $\rho$. For questions with only 2 answer options, Spearman rank correlation is unavailable.}
\label{table:full_safe}
\end{table*}

\begin{table*}
\centering
\footnotesize
\begin{tabular}{lcrrrr}
\toprule
\multicolumn{6}{c}{\textbf{Enticing}} \\
\textbf{Question} & \textit{n} & $r$ & $p$ of $r$ & $\rho$ & $p$ of $\rho$ \\ \midrule 
Medical Costs & 810 & 0.193 & 3.26e-08 & -- & -- \\
Animal Research & 770 & 0.074 & 4.07e-02 & -- & -- \\
Organic Foods & 829 & 0.089 & 1.06e-02 & 0.088 & 1.10e-02 \\
Gene Risks & 794 & 0.051 & 1.48e-01 & 0.051 & 1.53e-01 \\
Gene Disease & 771 & 0.109 & 2.43e-03 & -- & -- \\
Meat Hormone & 817 & 0.078 & 2.54e-02 & 0.071 & 4.25e-02 \\
New Treatments & 826 & 0.028 & 4.20e-01 & 0.04 & 2.50e-01 \\
Science Funding & 804 & 0.078 & 2.75e-02 & -- & -- \\
Food Additives & 820 & 0.054 & 1.25e-01 & -- & -- \\
Evolution & 830 & -0.03 & 3.87e-01 & -- & -- \\
Govt Retirement & 830 & -0.08 & 2.10e-02 & -- & -- \\
Church Econ & 809 & 0.088 & 1.24e-02 & 0.087 & 1.30e-02 \\
Immigrant Econ & 799 & -0.113 & 1.37e-03 & -0.123 & 4.72e-04 \\
Gas Prices & 825 & 0.03 & 3.94e-01 & 0.022 & 5.33e-01 \\
House Prices & 826 & 0.027 & 4.45e-01 & 0.037 & 2.93e-01 \\
Job Confidence & 827 & 0.049 & 1.60e-01 & 0.06 & 8.25e-02 \\
Race Econ & 819 & 0.026 & 4.59e-01 & 0.018 & 6.04e-01 \\
Corporate Econ & 810 & -0.071 & 4.31e-02 & -0.065 & 6.24e-02 \\
Benefits Econ & 822 & -0.013 & 7.06e-01 & -0.006 & 8.59e-01 \\
Antitrust Econ & 820 & -0.073 & 3.58e-02 & -0.074 & 3.44e-02 \\
Population & 815 & -0.083 & 1.75e-02 & -0.087 & 1.29e-02 \\
Energy Crisis & 802 & 0.019 & 6.00e-01 & 0.031 & 3.75e-01 \\
Public Ed. & 771 & 0.19 & 1.10e-07 & -- & -- \\
Child Standard & 769 & 0.225 & 2.71e-10 & -- & -- \\
China vs US & 799 & -0.124 & 4.34e-04 & -0.111 & 1.64e-03 \\
Race Relations & 788 & -0.216 & 9.41e-10 & -- & -- \\
Climate Change & 828 & 0.05 & 1.47e-01 & 0.074 & 3.43e-02 \\
Alzheimer Cure & 820 & 0.149 & 1.73e-05 & 0.156 & 6.98e-06 \\
Military Cost & 825 & -0.048 & 1.67e-01 & -0.048 & 1.68e-01 \\
Religion & 809 & -0.076 & 3.00e-02 & -- & -- \\
\bottomrule
\end{tabular}
\caption{Correlations of Pew Opinion responses with Enticing primal. Table shows number of responses \textit{n}, Pearson correlation coefficient $r$ with p-value $p$ of $r$, Spearman rank correlation coefficient $\rho$ with p-value $p$ of $\rho$. For questions with only 2 answer options, Spearman rank correlation is unavailable.}
\label{table:full_enticing}
\end{table*}

\begin{table*}
\centering
\footnotesize
\begin{tabular}{lcrrrr}
\toprule
\multicolumn{6}{c}{\textbf{Alive}} \\
\textbf{Question} & \textit{n} & $r$ & $p$ of $r$ & $\rho$ & $p$ of $\rho$ \\ \midrule 
Medical Costs & 765 & -0.022 & 5.50e-01 & -- & -- \\
Animal Research & 728 & -0.044 & 2.31e-01 & -- & -- \\
Organic Foods & 780 & 0.018 & 6.22e-01 & 0.023 & 5.28e-01 \\
Gene Risks & 748 & 0.047 & 1.96e-01 & 0.042 & 2.56e-01 \\
Gene Disease & 731 & -0.101 & 6.35e-03 & -- & -- \\
Meat Hormone & 768 & 0.13 & 3.05e-04 & 0.144 & 5.97e-05 \\
New Treatments & 776 & 0.129 & 3.04e-04 & 0.131 & 2.62e-04 \\
Science Funding & 759 & -0.157 & 1.46e-05 & -- & -- \\
Food Additives & 772 & 0.016 & 6.50e-01 & -- & -- \\
Evolution & 780 & 0.322 & 2.71e-20 & -- & -- \\
Govt Retirement & 780 & -0.104 & 3.60e-03 & -- & -- \\
Church Econ & 764 & 0.074 & 4.14e-02 & 0.08 & 2.72e-02 \\
Immigrant Econ & 754 & 0.108 & 3.01e-03 & 0.109 & 2.76e-03 \\
Gas Prices & 776 & 0.31 & 9.08e-19 & 0.301 & 1.04e-17 \\
House Prices & 774 & 0.115 & 1.33e-03 & 0.123 & 6.24e-04 \\
Job Confidence & 775 & 0.026 & 4.65e-01 & 0.038 & 2.96e-01 \\
Race Econ & 770 & -0.145 & 5.54e-05 & -0.134 & 2.03e-04 \\
Corporate Econ & 762 & 0.049 & 1.75e-01 & 0.036 & 3.28e-01 \\
Benefits Econ & 770 & -0.117 & 1.14e-03 & -0.115 & 1.34e-03 \\
Antitrust Econ & 771 & -0.106 & 3.08e-03 & -0.103 & 4.29e-03 \\
Population & 766 & -0.043 & 2.34e-01 & -0.053 & 1.45e-01 \\
Energy Crisis & 759 & -0.012 & 7.43e-01 & -0.008 & 8.35e-01 \\
Public Ed. & 737 & 0.136 & 2.12e-04 & -- & -- \\
Child Standard & 735 & 0.138 & 1.78e-04 & -- & -- \\
China vs US & 752 & -0.1 & 6.21e-03 & -0.108 & 2.99e-03 \\
Race Relations & 750 & -0.075 & 3.88e-02 & -- & -- \\
Climate Change & 778 & -0.202 & 1.23e-08 & -0.198 & 2.63e-08 \\
Alzheimer Cure & 771 & 0.15 & 2.95e-05 & 0.143 & 6.56e-05 \\
Military Cost & 774 & -0.113 & 1.60e-03 & -0.098 & 6.62e-03 \\
Religion & 759 & -0.145 & 6.02e-05 & -- & -- \\
\bottomrule
\end{tabular}
\caption{Correlations of Pew Opinion responses with Alive primal. Table shows number of responses \textit{n}, Pearson correlation coefficient $r$ with p-value $p$ of $r$, Spearman rank correlation coefficient $\rho$ with p-value $p$ of $\rho$. For questions with only 2 answer options, Spearman rank correlation is unavailable.}
\label{table:full_alive}
\end{table*}
\clearpage

\section{PI-18 Primal World Belief Inventory}
\label{sec:app-primals}
\begin{table*}
\centering
\resizebox{\linewidth}{!}{
\begin{tabular}{cl}
\toprule
{\bf Code} & {\bf Statement} \\ \hline
ed1 & In life, there’s way more beauty than ugliness. \\
am1 & It often feels like events are happening in order to help me in some way. \\
sd1 & I tend to see the world as pretty safe.\\
am2 & What happens in the world is meant to happen.\\
ed2x & While some things are worth checking out or exploring further, most things probably aren’t worth the effort.\\
ed3x & Most things in life are kind of boring.\\
ed4 & The world is an abundant place with tons and tons to offer.\\
ed5 & No matter where we are or what the topic might be, the world is fascinating.\\
ed6x & The world is a somewhat dull place where plenty of things are not that interesting.\\
sd2x & On the whole, the world is a dangerous place.\\
sd3x & Instead of being cooperative, the world is a cut-throat and competitive place.\\
am3x & Events seem to lack any cosmic or bigger purpose.\\
sd4x & Most things have a habit of getting worse.\\
am4 & The universe needs me for something important.\\
sd5 & Most things in the world are good.\\
am5 & Everything happens for a reason and on purpose.\\
sd6 & Most things and situations are harmless and totally safe.\\
ed7 & No matter where we are, incredible beauty is always around us.\\
\bottomrule
\end{tabular}
}
\caption{The 18 item Primal World Belief Inventory (PI-18). Response options are on a six point 0-5 scale: (5) Strongly agree, (4) Agree, (3) Slightly Agree, (2) Slightly Disagree, (1) Disagree, and (0) Strongly disagree. Items whose codes include ``x'' are reverse scored.}
\label{tab:primal_questions}
\end{table*}

\begin{table*}
\centering
\resizebox{\linewidth}{!}{
\begin{tabular}{cl}
\toprule
{\bf Primal} & {\bf Equation} \\ \hline
Good & $(sd1+sd2x+sd3x+sd4x+sd5+sd6+ed1+ed2x+ed3x+ed4+ed5+ed6x+
ed7+am1+am4)/15$ \\
Safe & $(sd1+sd2x+sd3x+sd4x+sd5+sd6)/6$ \\
Enticing & $(ed1+ed2x+ed3x+ed4+ed5+ed6x+ed7)/7$ \\
Alive & $(am1+am2+am3x+am4+am5)/5$ \\
\bottomrule
\end{tabular}
}
\caption{Equations for calculating high-level Primal scores from survey responses. }
\label{tab:primal-equations}
\end{table*}

The PI-18 consists of 18 multiple choice questions which assess worldview. Table~\ref{tab:primal_questions} shows the exact statements used in this survey. Participants rate their agreement with each statement on a scale from ``Strongly Agree'' to ``Strongly Disagree''. The responses are converted to high-level scores for each Primal using the equations in 
Table~\ref{tab:primal-equations}.

\section{Opinion prediction task details}
\label{sec:app-opiniontask}
Let $S$ and $T$ denote the set of seed and test multiple-choice opinion questions respectively. Response options are denoted with letters, and all questions include a "Prefer not to answer" option as required by our internal review board. 

In the {\it All Topics} setting $\mid S \mid~ = 9$ and $\mid T \mid~ = 21$ across 3 topics. In the {\it Cross Topic} setting $\mid S \mid~ = 10$ from Wave 34 and $\mid T \mid~ = 20$ from Waves 41 and 45. For a given user $u$, let $S^u = \{(a^u_i,e^u_i) | q_i \in S\}$ denote the user's responses $a$ and explanations $e$ for questions in $S$, and $T^u = \{a_j | q_j \in T\}$ be the user's responses for questions in $T$.

We predict unseen user responses for opinion questions by prompting and off-the-shelf language model $LM_\mathcal{T}$ with task-specific instructions $\mathcal{T}$. Prediction is conditioned on a test question $q_j\in T$ and user representation $U$. 

\begin{equation}
    \hat{a_j} = LM_\mathcal{T}(q_j,U)
\end{equation}

We restrict $\hat{a}$ to a single output token. Manual observations and experiments with longer outputs have shown that the models studied here produce a valid answer choice. The model output is correct when $\hat{a_j}=a^u_j$. Accuracy is the percentage of correct answer across all $T^u$ for users in the test split. 

\section{Model prompts and instructions}
\begin{figure}[!h]
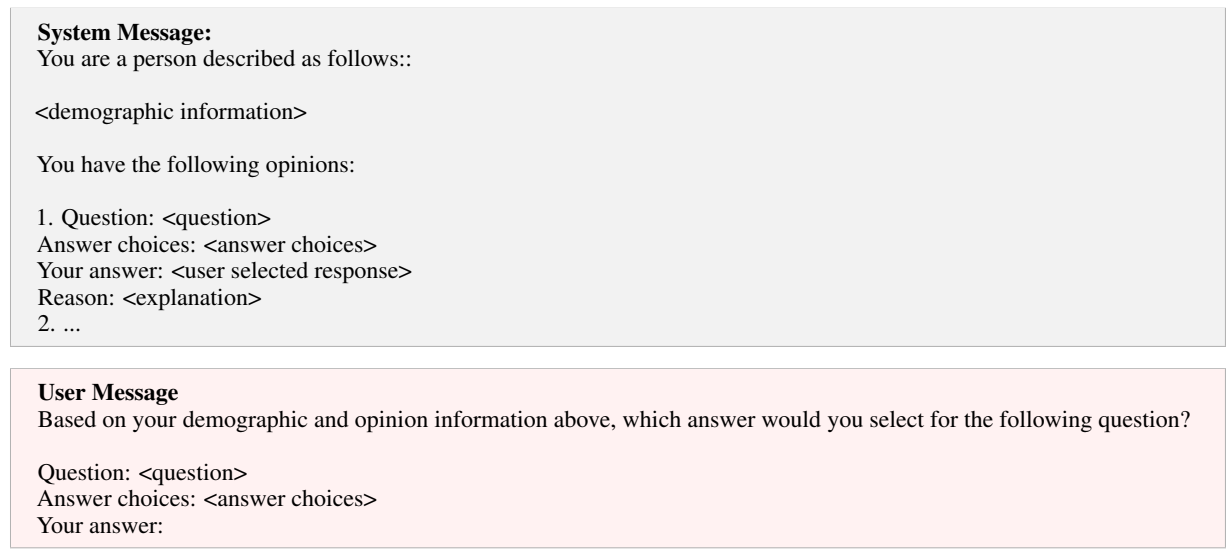

\small

\begin{mdframed}[linecolor=black!30,backgroundcolor=black!5]
\textbf{System Message:}

You are a person described as follows:: \\
\\
<demographic information>\\
\\
You have the following opinions:\\
\\
1. Question: <question>\\
Answer choices: <answer choices>\\
Your answer: <user selected response>\\
Reason: <explanation>\\
2. ...
\end{mdframed}

\begin{mdframed}[linecolor=black!30,backgroundcolor=red!5]
\textbf{User Message}\\
Based on your demographic and opinion information above, which answer would you select for the following question?\\
\\
Question: <question>\\
Answer choices: <answer choices>\\
Your answer:
\end{mdframed}

\caption{General prompt template for opinion prediction. Settings without demographics, opinions, or reasons omit these fields.}
\label{fig:prompt}
\end{figure}

\begin{figure}[!h]
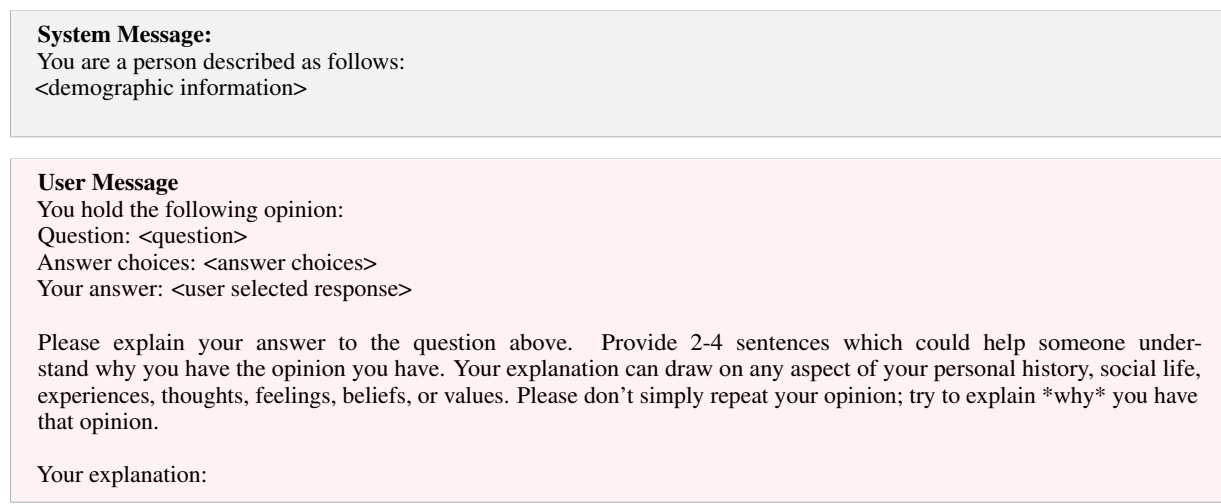

\small

\begin{mdframed}[linecolor=black!30,backgroundcolor=black!5]
\textbf{System Message:}

You are a person described as follows: \\
<demographic information>\\

\end{mdframed}

\begin{mdframed}[linecolor=black!30,backgroundcolor=red!5]
\textbf{User Message}\\
You hold the following opinion:\\
Question: <question>\\
Answer choices: <answer choices>\\
Your answer: <user selected response>\\
\\\
Please explain your answer to the question above. Provide 2-4 sentences which
could help someone understand why you
have the opinion you have. Your explanation can draw on any aspect of
your personal history, social life, experiences, thoughts, feelings, beliefs, or
values. Please don’t simply repeat your
opinion; try to explain *why* you have
that opinion.\\
\\
Your explanation:
\end{mdframed}

\caption{Prompt for generating \textsc{Finetuned} explanations, which is the same as the prompt given to survey respondents.}
\label{fig:prompt_finetuned}
\end{figure}

Figure~\ref{fig:prompt} shows the general prompt template for the opinion prediction experiments. For generating synthetic explanations from the \textsc{Finetuned} model, the prompt in Figure~\ref{fig:prompt_finetuned} is used. 

\section{Model Configuration}
Prediction experiments were conducted via API calls. Each model processed somewhere in the range of 500-750M tokens for these experiments. Hyper-parameters ``temperature$=0$'' and ``max\_tokens$=1$'' were used in the final results. We explored other max\_tokens settings $\in\{1,2,10\}$ to ensure this parameter wasn't impacting model outputs.  

The \textsc{Finetuned} opinion predictor was GPT-4o finetuned via OpenAI API on 174,399 tokens. The 20 most {\it helpful} explanations for each question from the training set are used as training data for this model (see Section~\ref{sec:helpful} for details on how helpfulness is calculated). The \textsc{Trained} Primals predictor was finetuned on 24,920,748 tokens. 

Explanations helpfulness calculations were done with Mistral 7B Instruct v0.3 on 8 A100 40GB GPU and took less than 24 hours. 

\section{Contextual information for Primal scores}
\begin{figure}[!h]
\small
\centering
\begin{mdframed}[linecolor=black!30,backgroundcolor=red!5]
\textbf{Example Primal Analysis Representation}\\

Good World Belief (vs. bad)

This primal concerns arguably the single most basic question anyone could ever ask: Is the world a good place?

What Does Good Predict?

When people are in places they see as really bad (or really good) we would expect them to behave in certain ways. When people see the world as a really bad place (or a really good place), we see a similar pattern.
Good is highly related to optimism, life satisfaction, curiosity, agreeableness, and virtually all aspects of psychological well-being—even subjectively reported physical health. Likewise, scoring low on Good (i.e., seeing the world as a bad place) is a risk factor for depression and many other mental health issues.
Though we don’t yet know where primals come from, we can definitely rule one thing out: seeing the world as Good is not a product of our material circumstances. Income and Good are related, but the relationship is very small. This small relationship may also be explained by Good causing someone to experience more professional success, instead of professional success causing one to see the world as Good.

You have a Good World Belief score of 3.0, which corresponds to seeing an OK world	The average American falls at 2.9. Those in this range are the most flexible in how they interpret ambiguity and see both pessimism and optimism as rational and appropriate -- different strokes. However, even within this range, where you fall may matter a lot. Because scores may vary considerably on Safe, Enticing, and Alive, different profiles suggest quite different approaches to life. For example, high Enticing and low Safe could make for "neurotic explorers;" low Enticing high Safe could make for "satisfied provincials," though both may score equally on Good.
\end{mdframed}
\begin{mdframed}[linecolor=black!30,backgroundcolor=red!5]

Safe World Belief (vs. dangerous)

This primal concerns to what degree we see the world as generally a safe place where threats are often overblown or a dangerous place where major threats really are everywhere. In their foundational 2019 research article, Clifton and the research team describe Safe in the following way:
Those low on Safe see a Hobbesian world defined by misery, decay, scarcity, brutality, and dangers of all sorts. Base rates for hazards—from germs to terrorism to getting stabbed in the back—are generally higher. In response to chronic external threats, they remain on high alert, often viewing the non-vigilant as irresponsible. Those high on Safe see a world of cooperation, comfort, stability, and few threats. To them, things are safe until proven otherwise, vigilance appears neurotic, risk is not that risky, and, in general, people should calm down.

What Does Safe Predict?

When people are in places they see as really dangerous, we would expect them to behave in certain ways, such as being irritable, tense, and so forth. When it comes to seeing the world as safe or dangerous, we see a similar pattern.
Safe is highly related to a host of personality and well-being variables. These include agreeableness, extraversion, interpersonal trust, and life satisfaction. Likewise, scoring low on Safe is related to neuroticism, depression, and loneliness.
Though we don’t yet know where primals come from, we are starting to realize that Safe scores are not simply determined by our environment. They appear more like a lens through which we view the world than a direct result of our experiences. For example, though people living in more affluent households are objectively safer than those living in poorer households because they can absorb various catastrophes, large medical bills, and so forth, there is no relationship between household income and Safe. Also, even though a case could be made that the world is a safer place for men than women, gender is completely unrelated to Safe.
Family members, friends, and colleagues, who score differently on Safe may have different baseline assumptions for vague situations. This may lead to a variety of misunderstandings and differing opinions. For example, those scoring high on Safe may find it easier to relax, be less suspicious of others, and assume nothing that bad will happen. Those scoring low on Safe may find it harder to relax, be more suspicious of others, and assume the worst.

You have a Safe World Belief score of 2.0, which corresponds to seeing a somewhat dangerous world	The average American scores a 2.5. Those in the middle are least likely to see themselves as holding a belief about the world. They can relate to everyone but also may be a bit baffled by the behavior of those on both extremes. Even within this large group (50\% of the population), how one scores may still matter a great deal.
\end{mdframed}

\caption{An example user representation with contextualizing information about Primal Beliefs and the user's score for each belief.}
\end{figure}
\medskip
\begin{figure}\ContinuedFloat
\small
\centering
\begin{mdframed}[linecolor=black!30,backgroundcolor=red!5]

Alive World Belief (vs. mechanistic)

This primal is about how much we see the world as full of intention and purpose that is interacting with us and wants our help. In their foundational 2019 paper, Clifton and his research team describe Alive like this:
Those low on Alive inhabit inanimate, mechanical worlds without awareness or intent. Since the universe never sends messages, it makes no sense to try to hear any. Those high on Alive sense that everything happens for a purpose and are thus sensitive to those purposes. To them, life is a relationship with an active universe that animates events, works via synchronicity, communicates, and wants help on important tasks.

What Does Alive Predict?

When people are in places thought to be overseen by something alive, we would expect them to behave in certain ways, such as trying to determine its thoughts and desires, seeking to interact with it, being responsive to perceived requests, and so forth. When it comes to seeing the world as alive, we see a similar pattern.
Alive is related to a variety of personality and well-being variables, for example, finding meaning in life, optimism, and life satisfaction. Also closely related to Alive are a tendency to have transcendent experiences, being a spiritual person, being more religious, and even being more extroverted. Likewise, scoring low on Alive is related to depression and ingratitude.
However, compared to Safe and Enticing, Alive is less related to personality and well-being variables while still playing an important role. For example, gratitude is very strongly related to Enticing. However, there is also a small but important unique relationship to Alive. In other words, to score high in gratitude, it’s critical to see the world as full of things to be grateful for, but also helps to have someone to be grateful to.
Interestingly, though Alive is strongly related to being religious, lots of non-religious people, even atheists, see the world as Alive. Within religious groups, there is also a lot of variation.
Family members, friends, and colleagues, who score differently on Alive may have different baseline assumptions for vague situations. This may lead to a variety of misunderstandings and differing opinions. For example, those scoring high on Alive may be more likely to read intentionality, meaning, and purpose into events that others perceive are not there. But they will also enjoy some mental health benefits that elude those who insist on seeing the world, and events that happen within the world, as mechanical and indifferent to them.

You have a Alive World Belief score of 2.8, which corresponds to seeing a not quite alive world	The average American scores a 2.8. This means they tend to see the world as slightly Alive, but not by much. Those in the middle are least likely to see themselves as holding a belief about the world at all, but they do. They can relate to everyone but may be a bit baffled by the behavior of those on both extremes. Even within this large group (50\% of the population), where one scores may still matter a great deal. These folks see the world as often animate, but often not. The universe has desires and intentions, but they don’t necessarily influence lots of events.

\end{mdframed}
\begin{mdframed}[linecolor=black!30,backgroundcolor=red!5]

Enticing World Belief (vs. dull)

This primal concerns how much we see the world as generally full of interesting and beautiful things or dull and ugly things. In their foundational 2019 paper, Clifton and his research team describe Enticing like this:
Those low on Enticing inhabit dull and ugly worlds where exploration offers low return on investment. They know real treasure—truly beautiful and fascinating things—is rare and treasure-hunting appropriate only when it’s a sure bet. Those high on Enticing inhabit an irresistibly fascinating reality. They know treasure is around every corner, in every person, under every rock, and beauty permeates all. Thus, life is a gift, boredom a misinformed lifestyle choice, and exploration and appreciation is the only rational way to live.

What Does Enticing Predict?

When people are in places they see as really enticing, we would expect them to behave in certain ways, such as being grateful, curious, open to experience, extroverted, and so forth. When it comes to seeing the world as enticing, we see a similar pattern.
Enticing is highly related to a host of personality and well-being variables. These include curiosity, gratitude, openness to experience, agreeableness, extroversion, engagement with life, positive emotion, meaning in life, a sense of accomplishment, and even having friends. Likewise, scoring low on Enticing is related to psychopathy, depression, and not trying.
Identifying Enticing is the single most important discovery of primals research so far. While Safe has been studied to a small degree, Enticing is both wholly unexplored and of immense practical importance. Enticing may play an enormous role in one’s own well-being and mental health; it’s just as important as Safe. Often Safe and Enticing correlate (increase and decrease together). After all, they are the two main reasons for seeing the world as Good (Alive is icing on the cake). However, sometimes Safe and Enticing work very differently. For example, it appears almost impossible to be a curious person or a grateful person without scoring high on Enticing. But, when it comes to those traits, Safe scores don’t matter much.
Family members, friends, and colleagues, who score differently on Enticing may have different baseline assumptions for vague situations. This may lead to a variety of misunderstandings and differing opinions. For example, those scoring high on Enticing may be able to spot amazing opportunities, but they may also be sucked in by bad ones. Those scoring low on Enticing may be less likely to be fooled or waste their time, but they may be more likely to miss opportunities and have life pass them by.

You have a Enticing World Belief score of 4.0, which corresponds to seeing a somewhat interesting world	The average American scores a 3.7 and sees the world as somewhat interesting. Those in the middle are least likely to see themselves as holding a belief about the world. They can relate to everyone but may be a bit baffled by the behavior of those on both extremes. Even within this large group (50\% of the population), where one scores may still matter a great deal.
\end{mdframed}

\caption{An example user representation with contextualizing information about Primal Beliefs and the user's score for each belief (cont.)}
\label{fig:analysis_rep}
\end{figure}

Figure~\ref{fig:analysis_rep} shows the contextual information included for GPT-4o models in the \textit{ +Primals} and \textsc{\dataset persona} user representations. This information is taken from \cite{clifton2019primalworldbeliefs}.

\clearpage
\section{Utility Computation}
\label{sec:app_utility}
Let $S$, $S^u$, $T$, and $T^u$ be as defined in Appendix~\ref{sec:app-opiniontask}.
We compute the utility of an explanation $e^u_i \in S^u$ to a LM by first establishing a baseline performance for the LM given a simple user representation $U = D+(q_i,a^u_i)$ for user demographics $D$. Here $(q_i,a^u_i)$ are the seed question and user response which the user has explained with explanation $e^u_i$. The baseline for computation is the expected log-probability assigned to $T^u$ by the LM conditioned on $U$:

\begin{align*}
     \mathbb{E}_{q_j \in T, a^u_j\in T^u}\mathcal{P}(a^u_j|q_j,U)
\end{align*}

We model $\mathcal{P}$ using the Mistral-7B-Instruct model~\citep{Jiang2023Mistral7} which we prompt with the user representation as well as the test question and answer choices. The answer choices are enumerated with letters; $\mathcal{P}$ is restricted to the letters corresponding to valid answer choices and renormalized. We consider $\mathcal{P}(a^u_j|\cdot)$ to represent the probability assigned by the language model to the user’s true choice.

The utility of $e_i^u$ is defined as the expected log-likelihood gain of $a_i^u$ when $P$ is conditioned on $e_i^u$: 

\begin{align*}
    \mathcal{M}(e_i^u) = \displaystyle \mathop{\mathbb{E}_{q_j \in T, a^u_j\in T^u}} & [\log(\mathcal{P}(a^u_j|q_j,U+e^u_i)) \\ & - \log(\mathcal{P}(a^u_j|q_j,U))]
\end{align*}

 $\mathcal{M}(e^u_i)$ then represents the change in probability of the true user answers under the model when provided with the extra information in the user’s explanation, averaged over the test questions. $\mathcal{M}$ can be and often is negative, as some explanations provide information which causes the language model to move probability mass away from the user's answers.

\end{document}